\documentclass{article}

\usepackage{arxiv}

\usepackage[utf8]{inputenc} 
\usepackage[T1]{fontenc}    
\usepackage{hyperref}       
\usepackage{url}            
\usepackage{booktabs}       
\usepackage{amsfonts}       
\usepackage{nicefrac}       
\usepackage{microtype}      
\usepackage{lipsum}
\usepackage{authblk}
\usepackage{graphicx}
\usepackage{amsmath}
\usepackage{comment}
\usepackage{xcolor}
\usepackage{adjustbox}
\usepackage{booktabs}
\usepackage{subcaption}
\usepackage{float}
\usepackage{algorithmic}
\usepackage{caption}
\usepackage{cite}
\usepackage{amsmath,amssymb,amsfonts}
\usepackage{algorithmic}
\usepackage{graphicx}
\usepackage{textcomp}
\usepackage{xcolor}
\usepackage{siunitx}
\usepackage{balance}
\usepackage{url}

\title{Feature extraction for plant growth estimation}

\author[1,2,3]{\textbf{S.~A.~Ngorima}}
\author[1]{\textbf{A.~S.~J.~Helberg}}
\author[1,2,3]{\textbf{M.~H.~Davel}}

\affil[1]{Faculty of Engineering, North-West University, South Africa}
\affil[2]{Centre for Artificial Intelligence Research, South Africa}
\affil[3]{National Institute for Theoretical and Computational Sciences, South Africa}
\affil[ ]{\texttt{aldringorima@gmail.com}}
\date{}
\date{\vspace{-2em}}
\begin{document}
\maketitle 
\begingroup
\renewcommand{\thefootnote}{}
\footnote{This version of the article has been accepted for publication, after peer review and is subject to Springer Nature’s AM terms of use, but is not the Version of Record and does not reflect post-acceptance improvements, or any corrections. The Version of Record is available online at: \url{https://doi.org/10.1007/978-3-032-11733-5_5}.}
\endgroup

\begin{abstract}
Precision agriculture requires the estimation of plant growth stages in real-time. When the plant growth stage is known, the wastage of resources in cultivation, such as nutrients and water, is reduced as only the required resources need to be supplied. Plants at different growth stages, however, have similar morphological features, which can make autonomous growth stage estimation difficult. This paper presents two feature extraction methods for growth stage estimation: one that uses a bank of Gabor filters and morphological operations, and the other that uses pre-trained convolutional neural networks (CNNs) and transfer learning. 
We test these methods on a publicly available plant growth stage dataset (``bccr-segset``) for two species, canola and radish, grown and captured under indoor conditions.  
The two proposed feature extraction methods are compared, using support vector machines and boosted trees as classifiers. 
We find that both methods are suitable for real-time applications, and that CNN features outperform the hand-crafted features, both with regard to speed and accuracy. 
The best system (VGG-19 features, classified with a radial basis function support vector machine) obtained an accuracy of 98.4\% for both species, processing an image in 0.08 seconds. 

\end{abstract}
\section{Introduction}
\label{intro}

Indoor farming systems, such as vertical farming, are well known for their ability to produce large yields while using fewer resources and taking up less space~\cite{van2021current}. In addition, they promote clean food production, as the use of chemicals such as herbicides is not typically required, since farming is done indoors and pests can often be controlled manually. 
Reducing the use of herbicides also reduces the number of farming resources required. Indoor farming also enables extensive data collection on plants, which is challenging to achieve in conventional farming.

Precision agriculture (PA), which involves the use of technology and analytical methods in agriculture, has significantly improved farming management and crop yields~\cite{logeshwaran2024improving,s24082647}. Precision agriculture in indoor environments utilises cameras, sensors, and actuators to identify and optimise the optimal ranges of the most critical environmental parameters for plant health and growth~\cite{s24082647}.  

Strengthening PA in indoor farming requires accurate plant growth stage estimates to identify the plant development stage. The plant development stage helps determine the appropriate nutrients for plants and it can also be used to predict yield~\cite{s21041406}. 
Applying only necessary resources to crops reduces farming costs. Broad-leaved plants in indoor agriculture grow quickly but are sensitive and require precise nutrition and controlled environments for high yield and quality.  
Good results are achieved when growth stage estimation is done regularly and preferably in real-time~\cite{alejandrino2020visual}. 
Several studies have been conducted on plant classification, achieving good results~\cite{s21041406,le2019effective,waldchen2018plant,le2020performances,logeshwaran2024improving}.  
To the best of our knowledge, the estimation of the growth stage of indoor-grown broad-leaved plants has not yet been thoroughly investigated.

We propose a computer vision-based approach for indoor farming that integrates feature extraction, morphological operations, and machine learning classification to estimate plant growth stages. We develop and compare two feature extraction methods as input to a real-time classifier. 
The first approach extracts features using Gabor filters, which mainly focus on textural features. The second approach uses CNN-based transfer learning for feature extraction, combined with classical machine learning classifiers. While transfer learning is not new to plant classification, its application for growth stage estimation remains limited. Side-by-side experiments are conducted and the results are compared using accuracy, confusion matrices, and execution time as metrics. 

The main contributions of this paper are: 
\begin{itemize}
\item We show that a pre-trained CNN model that uses transfer learning for plant feature extraction, concatenated with a classical machine learning algorithm, can yield accurate growth stage estimation, even when using a small training dataset. This method is fast and capable of predicting events in real-time.

\item Gabor filters are presented as a plant feature extractor that can also produce good classification results. We demonstrate that morphological operations can enhance the Gabor filter's feature extraction method, as they precisely highlight the shape and edges of an object within the frame. Manually designing a feature extraction method does not require large datasets.

\end{itemize}

The paper is organised as follows: Section~\ref{related works} reviews related work on plant detection and growth stage classification. Section~\ref{mm} describes the background and development of both the Gabor method and the transfer learning-based approach. In Section~\ref{mainD}, we describe the dataset, including its preprocessing and preparation to fit the task. Section~\ref{classification} outlines the classification algorithms (RBF-SVM and XGBoost) and their implementation. Finally, Section~\ref{results} presents the results, along with some additional analysis.

\section{Related Work}
\label{related works}

Machine learning techniques have transformed precision agriculture by increasing productivity while reducing costs~\cite{waqas2025applications,logeshwaran2024improving,s24082647}. Computer vision systems enable visual monitoring of plants, providing comprehensive insights into plant growth and health~\cite{mercier2025deep}. However, real-world applications face significant challenges: complex backgrounds, inconsistent lighting, and occlusion can make image segmentation and object recognition difficult~\cite{meraj2024computer}.

Plant classification typically relies on leaf features such as shape, colour, and texture. Waldchen et al.~\cite{waldchen2018plant} reviewed 120 studies and found that leaf shape and colour were the most widely used, with only 24 studies focusing on texture. This seems like a missed opportunity. The texture of the plant leaf changes significantly during the growth stages due to anatomical developments: changes in the epidermis, tissue reorganisation, and changes in the venation pattern. These changes may be most visible through texture variations, suggesting that texture-based approaches could be more discriminative for growth-stage classification. Gabor filters have shown superior performance in texture analysis in various applications~\cite{cope2010plant,salman2013pavement,vyas2016iris}. Their application to plant growth stage classification has not been explored. 

Recent growth stage classification studies have used various approaches. CNN-based methods have shown promise, with Murata et al.~\cite{murata2019study} achieving an accuracy of (89.3\%) for the estimation of the rice growth stage, and Rasti et al.~\cite{rasti2021crop} successfully identified 12 growth stages of wheat and 11 growth stages of barley. However, these approaches typically require large datasets and have a prolonged inference time~\cite{le2020performances}. Colour-based methods have also shown effectiveness, with the latest studies achieving 79-88\% accuracy for lettuce growth stage classification~\cite{concepcion2020lettuce}.

Transfer learning has emerged as an effective approach for plant classification tasks, adapting pre-trained CNN knowledge to new domains while reducing computational requirements and training data needs~\cite{logeshwaran2024improving}. Popular architectures include VGG~\cite{brusilovsky:simonyan2014very}, and ResNet~\cite{he2016deep}, typically pre-trained on ImageNet. 
However, whether features learnt from general object recognition translate effectively to the task of distinguishing consecutive growth stages, where morphological differences are often subtle, remains unclear.

\section{Method}
\label{mm}
This study proposes two feature extraction approaches for plant growth stage estimation: a Gabor filter method enhanced with morphological operations and a CNN transfer learning approach. The Gabor method targets textural features while leveraging morphological operators to highlight leaf contour, shape, and size characteristics. The transfer learning method uses pre-trained convolutional neural networks to extract plant features: specifically, the VGG-16, VGG-19, and ResNet-50 models, chosen for their demonstrated performance in various domains.

\subsection{Gabor Filter Method}
\label{gm}

A Gabor filter~\cite{gabor1946theory} combines a sinusoidal wave with a Gaussian envelope to analyse the texture of the image by locating the frequency content in specified directions. The Gaussian envelope captures spatial details, while the sinusoidal component identifies frequency-specific information~\cite{daugman1985image}. 
This combination appears particularly suited for plant analysis as it remains relatively stable under varying illumination and contrast conditions.

The filtering process involves convolving the image with multiple Gabor kernels at varying frequencies and orientations. The parameters of each kernel determine the type and scale of the features extracted. Specifically, the sinusoidal wave's frequency ($\lambda$) and orientation ($\theta$) define the type of texture captured, while the Gaussian envelope’s standard deviation ($\sigma$) and aspect ratio ($\gamma$) control the spatial extent and shape of the feature analysis region.
Mathematically, the general complex Gabor filter is defined as:

\begin{equation}
g(x, y, {\theta}, {\lambda}, {\psi}, {\sigma}, {\gamma}) = \exp \left( -\frac{x'^2+\gamma^2y'^2} {2\sigma^2}\right) \exp{\left(i\left(2\pi\frac{x'}{\lambda}+\psi\right)\right)}
\label{eqn:complex}
\end{equation}
with real and imaginary components given respectively by:
\begin{equation}
g_{R}(x, y, \theta, \lambda, \psi, \sigma, \gamma) = \exp \left( -\frac{x'^2+\gamma^2y'^2} {2\sigma^2}\right) \cos\left(2\pi\frac{x'}{\lambda}+\psi\right)
\label{eqn:real}
\end{equation}
\begin{equation}
g_{I}(x, y, \theta, \lambda, \psi, \sigma, \gamma) = \exp \left( -\frac{x'^2+\gamma^2y'^2} {2\sigma^2}\right) \sin\left(2\pi\frac{x'}{\lambda}+\psi\right)
\label{eqn:imaginary}
\end{equation}
where $x'=x\cos(\theta)+y\sin(\theta)$ and $y'=y\cos(\theta)-x\sin(\theta)$. The parameters control the filter behaviour: $\theta$ represents orientation, $\lambda$ defines wavelength, $\sigma$ sets the Gaussian standard deviation, $\psi$ adjusts phase offset, and $\gamma$ determines the spatial aspect ratio.

\paragraph{Parameter Selection and Implementation:}
We create a Gabor filter bank by empirically determining optimal parameter combinations through testing on representative training samples from each class. Parameter values are selected by examining PCA-t-SNE visualisations of extracted features and the monitoring validation score. This process is time-consuming but essential for capturing meaningful textural distinctions specific to plant images.

The phase offset had a minimal impact on classification accuracy, so we set $\psi = 0$ for simplicity. Table~\ref{table:g filter parametr} shows the final parameter selection after extensive experimentation. The filter bank uses four distinct frequencies across 8 orientations, generating 32 distinct response maps per image.

\begin{table}[ht!]
	\centering
	\caption{Gabor filter parameters.}
        \label{table:g filter parametr}
	\footnotesize
	\setlength{\tabcolsep}{5pt}
        \setlength{\extrarowheight}{5pt}
	\begin{tabular}{c c c c c c c}
		\hline
		 \centering\textbf{Filter} & \textbf{Kernel size} & \textbf{$\theta$} & \textbf{$\lambda$} & \textbf{$\psi$} & \textbf{$\sigma$} & \textbf{$\gamma$} \\
		\hline
		1 & $3\times3$ & $ \frac{\pi}{4}$ & $ \frac{\pi}{10}$ & 0 & 0.5 & 0.5 \\
		\hline
		2 & $3\times3$ & $ \frac{\pi}{4}$ & $ \frac{\pi}{8}$ & 0 & 0.5 & 0.5 \\
		\hline
		3 & $3\times3$ & $ \frac{\pi}{2}$ & $ \frac{\pi}{10}$ & 0 & 0.5 & 0.5 \\
		\hline
		4 & $3\times3$ & $ \frac{\pi}{2}$ & $ \frac{\pi}{8}$ & 0 & 0.5 & 0.5 \\
		\hline
	\end{tabular} 
\end{table}

The 3$\times$3 kernel size was retained while we extensively tuned other Gabor parameters. Rimiru et al.~\cite{rimiru2022gabornet} found that a 3$\times$3 Gabor kernel achieved 99.68\% efficiency on CIFAR-10, compared to 98.69\% and 98.46\% for 5$\times$5 and 7$\times$7 kernels, respectively, suggesting diminishing returns from larger kernels in their architecture. We adopt the 3$\times$3 kernel based on this evidence of diminishing returns from larger kernels and its computational efficiency for real-time performance.

\begin{figure}[ht!]
    \centering
    \includegraphics[width=5cm]{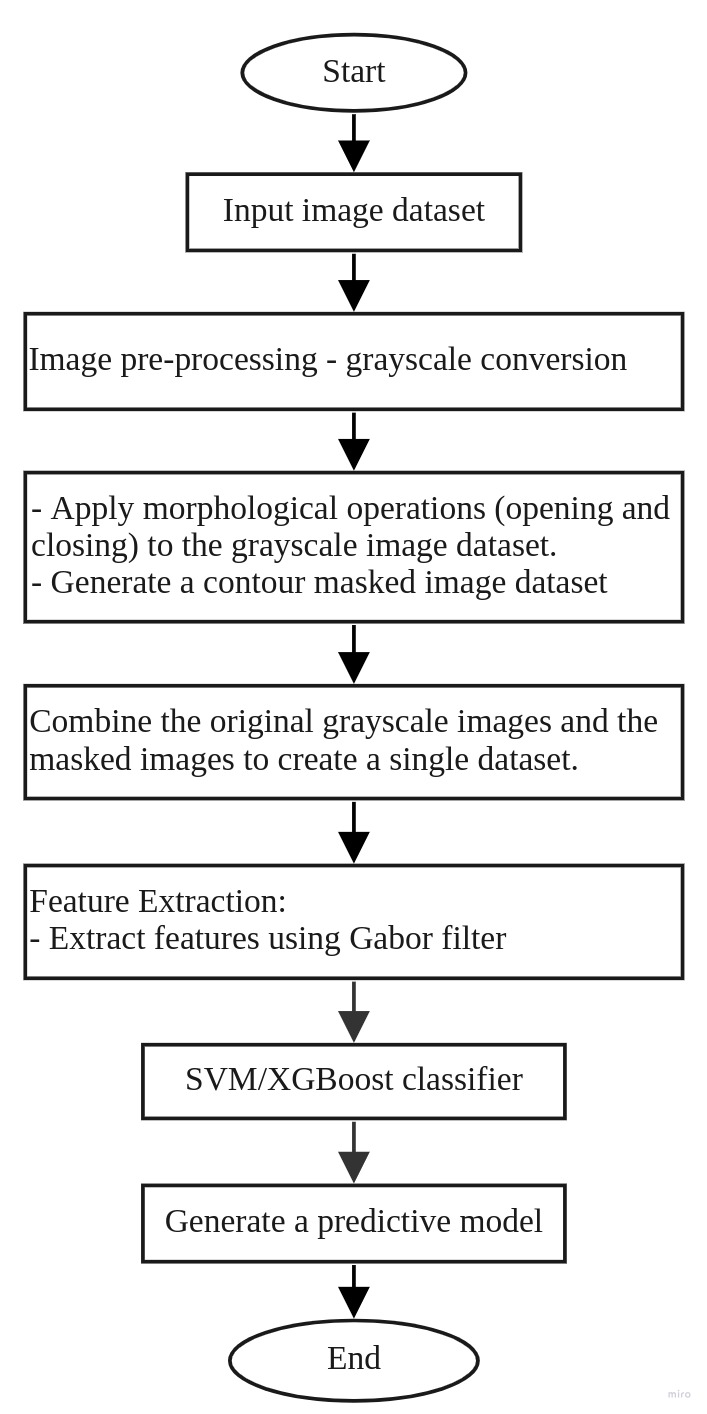}
    \caption{Gabor method flowchart with morphological operators.}
    \label{fig:galaxy1}
\end{figure}

\begin{figure}[ht!]
    \centering
    \begin{subfigure}[b]{0.28\textwidth}
        \includegraphics[width=\textwidth]{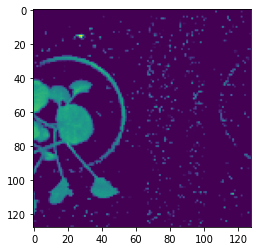}
        \caption{Original image}
        \label{fig:Original image}
    \end{subfigure}
    \hfill
    \begin{subfigure}[b]{0.28\textwidth}
        \includegraphics[width=\textwidth]{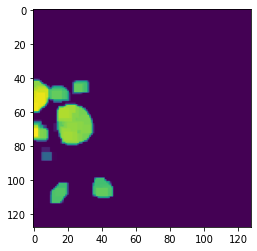}
        \caption{Opened image}
        \label{fig:Opened image}
    \end{subfigure}
    \hfill
    \begin{subfigure}[b]{0.28\textwidth}
        \includegraphics[width=\textwidth]{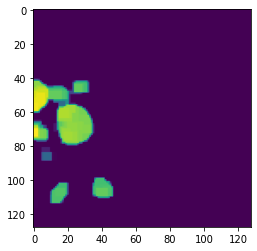}
        \caption{Closed image}
        \label{fig:Closed image}
    \end{subfigure}

    \par\medskip 

    \begin{subfigure}[b]{0.28\textwidth}
        \includegraphics[width=\textwidth]{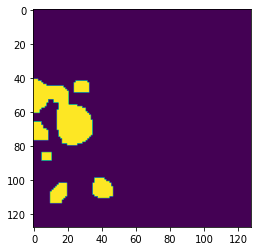}
        \caption{Thresholded image}
        \label{fig:Thresholded image}
    \end{subfigure}
    \hspace{0.06\textwidth} 
    \begin{subfigure}[b]{0.28\textwidth}
        \includegraphics[width=\textwidth]{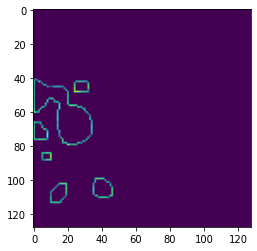}
        \caption{Masked image}
        \label{fig:Contour masked image}
    \end{subfigure}

    \caption{Morphological operations workflow: original to contour-masked image.}
    \label{fig:morpho operations}
\end{figure}

\paragraph{Morphological Operations and Feature Extraction Workflow:}
Figure~\ref{fig:galaxy1} shows the complete feature extraction workflow using Gabor filters combined with morphological preprocessing. Initially, images are resized to 128$\times$128 pixels and converted to greyscale. Morphological operations are then applied to highlight the contours and shape characteristics of plants. This preprocessing involves opening (noise removal), closing (object enhancement), thresholding (plant segmentation), and contour masking (shape isolation), as illustrated step-by-step in Figures~\ref{fig:Original image} through~\ref{fig:Contour masked image}. 

\begin{figure}[ht!]
    \centering
    \begin{subfigure}[b]{0.25\textwidth}
        \includegraphics[width=\textwidth]{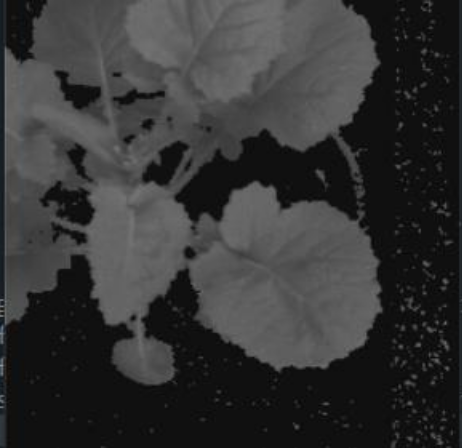}
        \caption{Original image}
        \label{fig:Original plant image}
    \end{subfigure}
    \hfill
    \begin{subfigure}[b]{0.25\textwidth}
        \includegraphics[width=\textwidth]{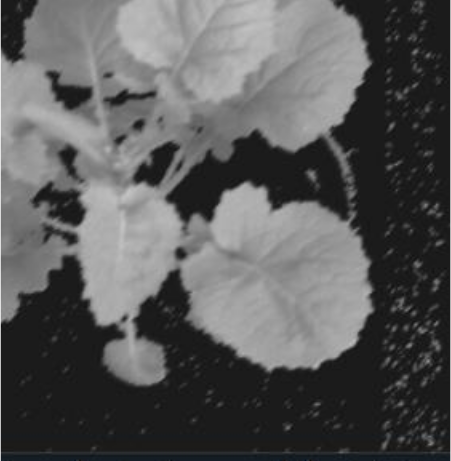}
        \caption{Filter 1 response}
        \label{fig:Filtered image by filter 1}
    \end{subfigure}
    \hfill
    \begin{subfigure}[b]{0.25\textwidth}
        \includegraphics[width=\textwidth]{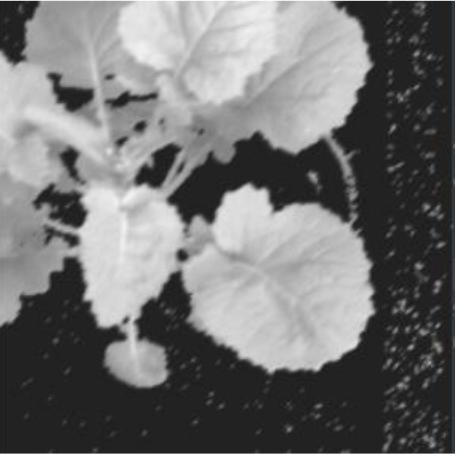}
        \caption{Filter 2 response}
        \label{fig:Image features by filter 2}
    \end{subfigure}

    \par\medskip 

    \begin{subfigure}[b]{0.25\textwidth}
        \includegraphics[width=\textwidth]{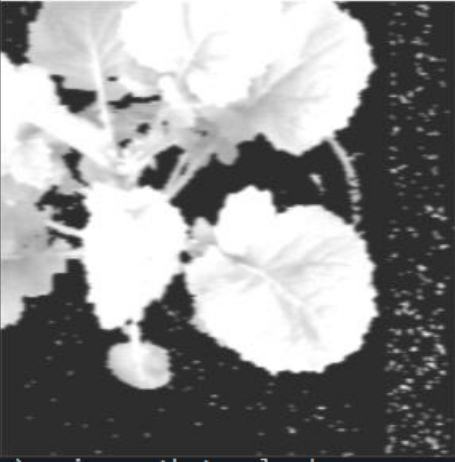}
        \caption{Filter 3 response}
        \label{fig:Image features by filter 3}
    \end{subfigure}
    \hspace{0.06\textwidth}
    \begin{subfigure}[b]{0.25\textwidth}
        \includegraphics[width=\textwidth]{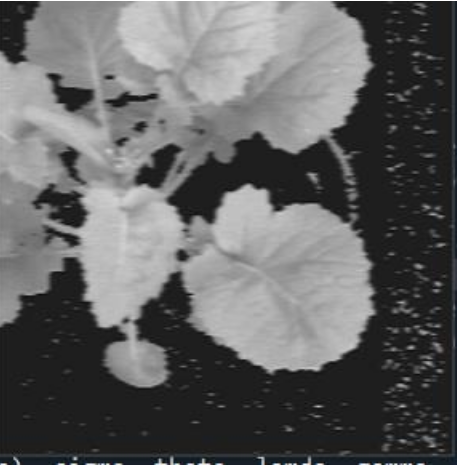}
        \caption{Filter 4 response}
        \label{fig:Image features by filter 4}
    \end{subfigure}

    \caption{Feature responses from the Gabor filter bank applied to a plant image.}
    \label{fig:Gaborexc}
\end{figure}

Both original greyscale and morphologically processed contour-masked images undergo convolution with the Gabor filter bank. This generates multiple feature response maps, as demonstrated in Figure~\ref{fig:Gaborexc}, where each filter response highlights specific textures and structural plant features. These feature maps subsequently form the input for the machine learning classifiers. 

\subsection{Transfer Learning Method}
\label{pre-trained}

Pre-trained CNNs offer an alternative feature extraction approach by leveraging knowledge gained from large-scale datasets. CNN architectures typically stack convolutional layers for feature extraction, pooling layers for dimensionality reduction, and fully-connected layers for classification~\cite{le2020performances}. 
We repurpose these networks by removing classification layers and using only the convolutional components as feature extractors. This approach seems promising because convolutional layers capture general visual elements, edges, colours, and textures, which may transfer well to plant analysis~\cite{mukti2019transfer}. 
Our implementation uses the pre-trained convolutional layers of VGG-16, VGG-19~\cite{brusilovsky:simonyan2014very}, and ResNet-50~\cite{he2016deep} without retraining these components. 

Images are first greyscaled and fed directly to these feature extractors. We deliberately avoid fully-connected layers since they contain most network parameters; around 123 million of VGG-16's 138 million parameters are found in classification layers~\cite{rezende2018malicious}. Replacing these computationally expensive components with classical machine learning algorithms is expected to improve prediction speed while maintaining feature quality, which aligns with our real-time processing goals.

\section{Dataset}
\label{mainD}
The dataset used in this work is the publicly available “bccr-segset”, originally compiled by Edith Cowan University, Australia, for crop and weed detection~\cite{le2019effective}. It contains approximately 30,000 images of canola, radish, and maize captured at four distinct growth stages under controlled indoor conditions, with images acquired at $228\times228$pixels from a top-mounted camera. For our study, we repurposed this data set for the classification of growth stages, focusing exclusively on broad-leafed plants (canola and radish). Maize images were excluded, and a separate “background” class was added to remove bias from empty frames. To address class imbalance, we used stratified sampling during dataset partitioning and stratified cross-validation for evaluation.


\subsection{Dataset partitioning}
\label{datapreprocesing}
We grouped images according to the pre-labelled growth stage and plant species. Each dataset was partitioned using stratified random sampling to maintain proportional class representation across all growth stages: approximately 90\% for training and 10\% for testing. This resulted in training sets of 4,831 images (canola) and 4,094 images (radish), with corresponding test sets of 547 and 525 images, respectively. 

During hyperparameter optimisation, we further split the training data using stratified sampling, reserving 15\% as a validation set for model selection while using the remaining 85\% for model training. Feature extraction was performed on all images, but hyperparameter tuning of the classification methods (RBF-SVM and XGBoost) was performed using 3-fold stratified cross-validation in the training set to prevent data leakage. The validation set guided hyperparameter selection, while the test set was held out and untouched during model development for final performance evaluation.
After selecting the optimal hyperparameters, each model was retrained on the entire training set and then evaluated once on the held-out test set.

The image input sizes were adjusted to fit the dimensionality requirements of the feature extractor methods proposed in this work. The input dimensionality of the CNN models was set to 224$\times$224$\times$1, while that of the Gabor method was set to 128$\times$128$\times$1. After preprocessing, the dataset sizes per class are summarised in Table~\ref{datasetdetails}.


\begin{table}[ht!]
	\centering
	\caption{The number of images in the two species datasets at different growth stages.}
        \label{datasetdetails}
	\footnotesize
	\begin{tabular}{c c c}
		\hline
		\centering \textbf{Class} & \textbf{Canola dataset} & \textbf{Radish dataset} \\
		\hline
		Background & 2387  &  799 \\
		\hline
		Growth stage 1 & 527 &  782 \\
		\hline
		Growth stage 2 & 360 & 780 \\
		\hline
		Growth stage 3 & 1629 & 822 \\
		\hline
		Growth stage 4 & 475 & 1436 \\
		\hline
		\textbf{Total} & \textbf{5378}  & \textbf{4619}  \\
		\hline
	\end{tabular} 
\end{table}

\subsection{Dataset visualisation}
We used principal component analysis (PCA) \cite{abdi2010principal} and t-SNE \cite{van2008visualizing} to visualise the structure and correlation between each of the growth stages per plant class in the dataset. t-SNE reduces data dimensionality such that data points closer together in the original high-dimensional space remain close in the two-dimensional representation. The dimensionality reduction process is entirely unsupervised; labels are only used to colour-code the nodes when plotting. PCA was used as a preprocessing step to reduce dimensions before applying t-SNE to visualise the growth stage clusters. A t-SNE visualisation indicates whether classes can be easily separated without the use of feature extraction methods. 

\begin{figure}[ht!]
    \centering
    \begin{subfigure}[b]{0.48\textwidth}
        \centering
        \includegraphics[width=\textwidth]{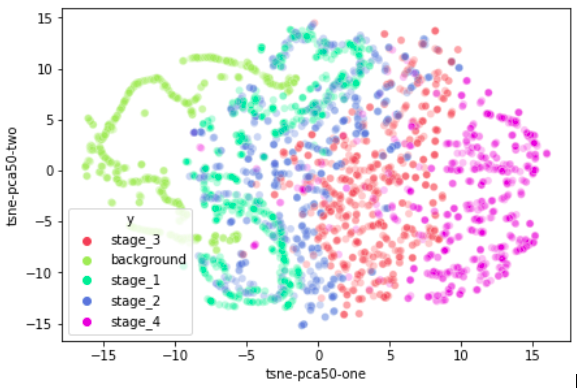}
        \caption{Canola dataset's five classes}
        \label{fig:galaxy5}
    \end{subfigure}
    \hfill
    \begin{subfigure}[b]{0.48\textwidth}
        \centering
        \includegraphics[width=\textwidth]{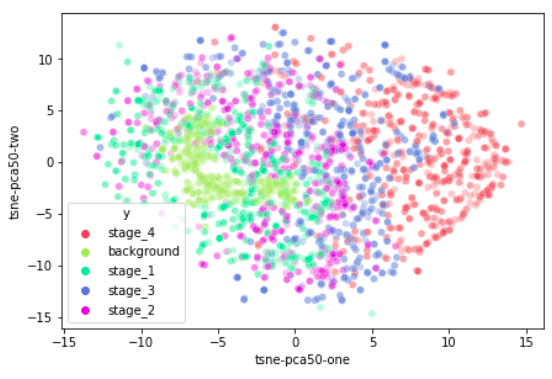}
        \caption{Radish dataset's five classes}
        \label{fig:galaxy6}
    \end{subfigure}
    \caption{Visualisation of the five classes for both datasets.}
    \label{fig:combined_datasets}
\end{figure}

Figure~\ref{fig:galaxy5} and Figure~\ref{fig:galaxy6} show the two-dimensional visualisation of the canola dataset and the radish dataset created from the ``bccr-segset'' dataset. The visualisations show that the background and growth stage 4 classes can be distinguished more clearly, but that the other classes, stages 1, 2, and 3, have overlapping patterns. Classifying these classes without identifying and extracting more nuanced features is expected to produce high misclassification rates. 

\section{Classification}
\label{classification}

We first describe the process to select classification methods (Section \ref{sec:method_selection}), before discussing the implementation and optimisation of each method (Section \ref{sec:implementation}).

\subsection{Method Selection}
\label{sec:method_selection}

After extracting high-level features via CNN or the Gabor pipeline, we selected SVMs and XGBoost over a standard multilayer perceptron (MLP) for classification due to the following technical considerations:

 \paragraph{Regularisation and Limited Data:} Our datasets were moderate in size, so overfitting was a concern. Fully connected neural layers (MLPs) introduce a large number of trainable parameters and tend to overfit when training data is very limited or noisy~\cite{pr13061825}. In contrast, SVMs and XGBoost have strong built-in regularisation (maximum-margin classification for SVM, and shrinkage/ensemble averaging for XGBoost).
 
 \paragraph{Inference Speed and Efficiency:} The SVM and XGBoost classifier ensures quick inference, since SVMs require only support vectors and kernel parameters, while XGBoost uses tree-based structures with inherently fewer parameters than dense neural network layers.  SVMs compute a dot product with support vectors, and XGBoost uses simple tree traversals; both operations are efficient. Wang et al.~\cite{pr13061825} demonstrated that using an SVM as the classification layer reduced inference latency by ~31\% 
 compared to a CNN with a softmax layer. 
 
 We chose SVM and XGBoost over other classical approaches, as SVMs are effective in high-dimensional feature spaces, such as those produced by CNN embeddings or Gabor filters, and incorporate explicit regularisation that improves generalisation~\cite{cortes1995support}. 
XGBoost was preferred because gradient boosting reduces bias by focusing on difficult-to-classify samples, often leading to improved accuracy and compact models while maintaining efficient inference~\cite{chen2016xgboost}.
%
 
\subsection{Implementation and Hyperparameter Optimisation}
\label{sec:implementation}

\textbf{RBF-SVM Implementation}
RBF-SVM maps non-linear input data into a higher-dimensional space where linear separation is possible. Two hyperparameters control the RBF-SVM's behaviour: the regularisation parameter $C$ (penalty for misclassification) and $\gamma$ (kernel spread). Higher $C$ values yield smaller margins, while larger $\gamma$ values create more flexible decision boundaries.
%

We optimised the RBF-SVM hyperparameters using scikit-learn’s RandomizedSearchCV with 3-fold stratified cross-validation on a representative training subset. The search ranges were $C \in \{2^{-2}, \dots, 2^6\}$ and $\gamma \in \{2^{-10}, \dots, 2^{-2}\}$. After 30 iterations, the optimal values were $C = 32$ and $\gamma = 0.00001$.
%
%

\textbf{XGBoost Implementation}
XGBoost employs ensemble learning by combining multiple decision trees, enhanced with L1 and L2 regularisation to improve generalisation. The algorithm handles missing values naturally and often performs well on structured data, characteristics that seemed appealing to our feature-based approach.
Following the same optimisation procedure as before, we tuned five key hyperparameters. Table~\ref{xgboostH_combined} presents both search spaces and optimal values discovered through RandomizedSearchCV.
\begin{table}[ht!]
	\centering
	\caption{XGBoost hyperparameter search space and optimal values.}
        \label{xgboostH_combined}
	\footnotesize
	\setlength{\tabcolsep}{5pt}
        \setlength{\extrarowheight}{5pt}
	\begin{tabular}{c c c}
		\hline
		\centering \textbf{Hyperparameter} & \textbf{Search Space} & \textbf{Optimum Value} \\
		\hline
		Learning rate & [0.2 to 0.7] & 0.5 \\
		\hline
		Max depth & [5 to 12] & 8 \\
		\hline
		Min child weight & [1 to 7] & 7 \\
		\hline
		gamma & [0.3 to 0.7] & 0.6 \\
		\hline
		Colsample bytree & [0.3 to 0.7] & 0.7 \\
		\hline
	\end{tabular} 
\end{table}

\section{Results and Discussion}
\label{results}
This section discusses the evaluation process and compares the results achieved by the two methods. We begin by visualising the extracted features using t-SNE and PCA for qualitative analysis, then conduct a quantitative comparison using accuracy, confusion matrices, and execution time as metrics.

\subsection{Feature Visualisation and Qualitative Analysis}
\begin{figure}[ht!]
    \centering
    \begin{subfigure}[b]{0.45\textwidth}
        \includegraphics[width=\textwidth]{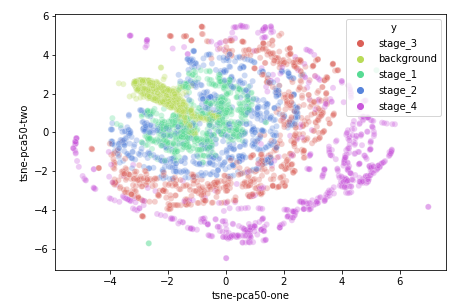}
        \caption{Gabor method}
        \label{fig:canola_gabor}
    \end{subfigure}
    \hspace{0.05\textwidth}
    \begin{subfigure}[b]{0.45\textwidth}
        \includegraphics[width=\textwidth]{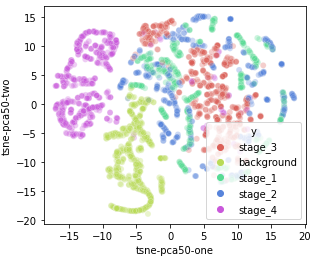}
        \caption{VGG-16 method}
        \label{fig:canola_vgg16}
    \end{subfigure}

    \par\medskip

    \begin{subfigure}[b]{0.45\textwidth}
        \includegraphics[width=\textwidth]{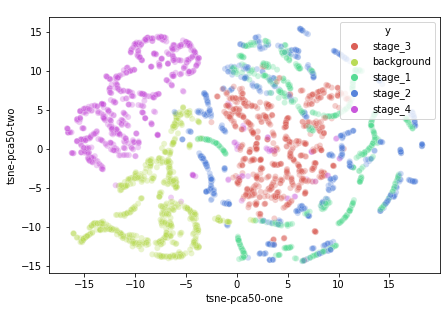}
        \caption{VGG-19 method}
        \label{fig:canola_vgg19}
    \end{subfigure}
    \hspace{0.05\textwidth}
    \begin{subfigure}[b]{0.45\textwidth}
        \includegraphics[width=\textwidth]{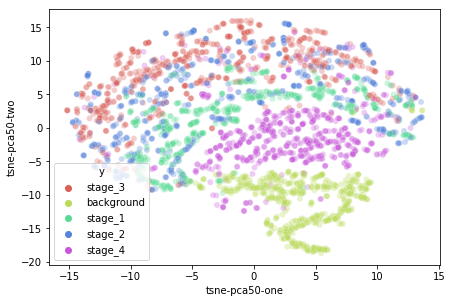}
        \caption{ResNet-50 method}
        \label{fig:canola_resnet}
    \end{subfigure}

    \caption{Feature distributions of the canola training dataset using Gabor, VGG-16, VGG-19, and ResNet-50 feature extraction methods.}
    \label{fig:canola_feature_dist}
\end{figure}

\begin{figure}[ht!]
    \centering
    \begin{subfigure}[b]{0.45\textwidth}
        \includegraphics[width=\textwidth]{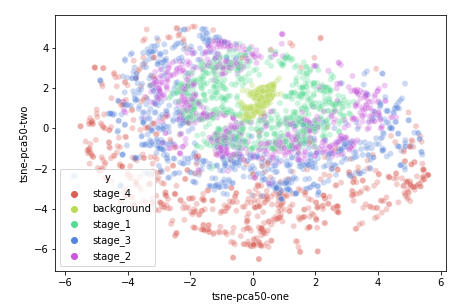}
        \caption{Gabor method}
        \label{fig:radish_gabor}
    \end{subfigure}
    \hspace{0.05\textwidth}
    \begin{subfigure}[b]{0.45\textwidth}
        \includegraphics[width=\textwidth]{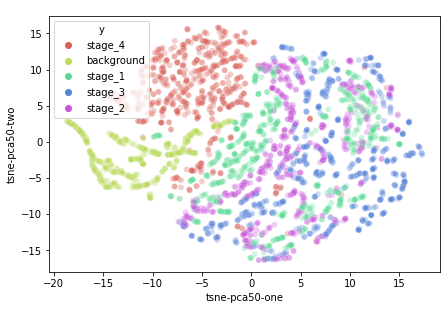}
        \caption{VGG-16 method}
        \label{fig:radish_vgg16}
    \end{subfigure}

    \par\medskip

    \begin{subfigure}[b]{0.45\textwidth}
        \includegraphics[width=\textwidth]{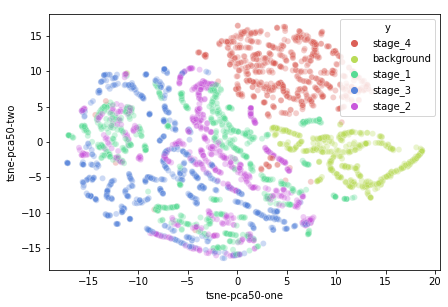}
        \caption{VGG-19 method}
        \label{fig:radish_vgg19}
    \end{subfigure}
    \hspace{0.05\textwidth}
    \begin{subfigure}[b]{0.45\textwidth}
        \includegraphics[width=\textwidth]{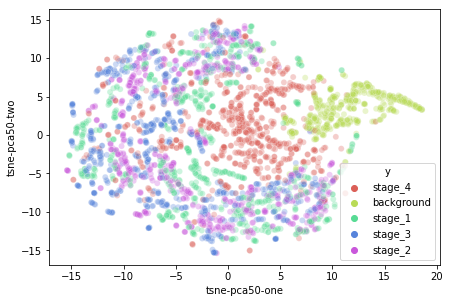}
        \caption{ResNet-50 method}
        \label{fig:radish_resnet}
    \end{subfigure}

    \caption{Feature distributions of the radish training dataset using Gabor, VGG-16, VGG-19, and ResNet-50 feature extraction methods.}
    \label{fig:radish_feature_dist}
\end{figure}

Figure~\ref{fig:canola_feature_dist} and Figure~\ref{fig:radish_feature_dist} show t-SNE-PCA visualisations of the different feature extraction approaches (Gabor filters, VGG-16, VGG-19, and ResNet-50) on both datasets, with each colour representing a different growth stage or background class. 

It is observed that VGG-16 and VGG-19 extract features that can be projected into separable spaces, while the Gabor method features also demonstrate reasonable separability. However, ResNet-50 showed lower clustering quality with more overlap between growth stages, likely because its residual connections, while effective for very deep networks, preserve low-level features that may act as noise rather than discriminative features for subtle plant texture variations. Despite this, all feature extraction methods appear capable of capturing features useful for growth stage classification, though some overlap remains visible across all techniques. This suggests that non-linear classifiers are necessary for effective classification.

\subsection{Classification Accuracy Evaluation}
\begin{figure}[ht!]
    \centering
    \begin{subfigure}[b]{0.45\textwidth}
        \centering
        \includegraphics[width=\textwidth]{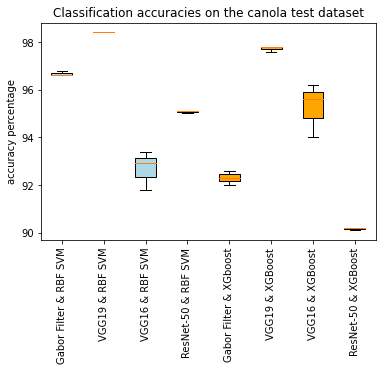}
        \caption{Canola test dataset}
        \label{fig:canola_accurcy}
    \end{subfigure}
    \hfill
    \begin{subfigure}[b]{0.45\textwidth}
        \centering
        \includegraphics[width=\textwidth]{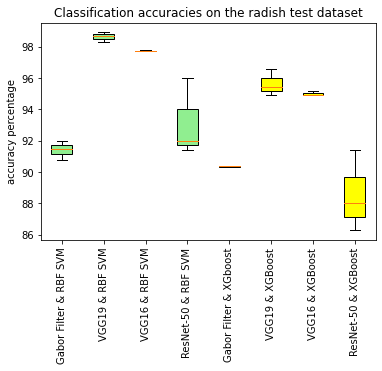}
        \caption{Radish test dataset}
        \label{fig:radish_acuracy}
    \end{subfigure}
    \caption{Classification accuracies on the test datasets.}
    \label{fig:combined_accuracies}
\end{figure}

We evaluated the classification accuracy using three-fold stratified cross-validation. Figure~\ref{fig:combined_accuracies} shows boxplots of classification accuracy on both test datasets for various model combinations.
VGG-19 concatenated with RBF-SVM achieved the highest accuracy of 98.4\% on both test datasets, demonstrating consistent performance across plant species. The Gabor filter method obtained competitive accuracies of 96.7\% on canola and 91.4\% on radish. ResNet-50 consistently achieved the lowest classification accuracy with both classification algorithms. The box plots also reveal the stability of the performance: VGG-19 showed consistent results across the folds, while other methods exhibited some variance.

\begin{figure}[ht!]
    \centering
    \begin{subfigure}[b]{0.45\textwidth}
        \includegraphics[width=\textwidth]{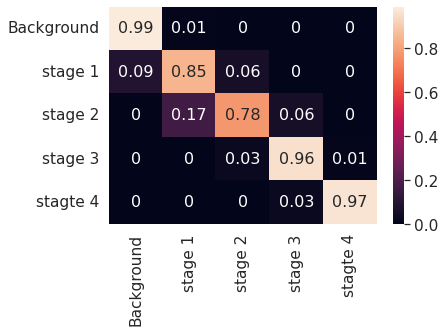}
        \caption{Gabor method}
        \label{fig:GaborR}
    \end{subfigure}
    \hspace{0.005\textwidth}
    \begin{subfigure}[b]{0.45\textwidth}
        \includegraphics[width=\textwidth]{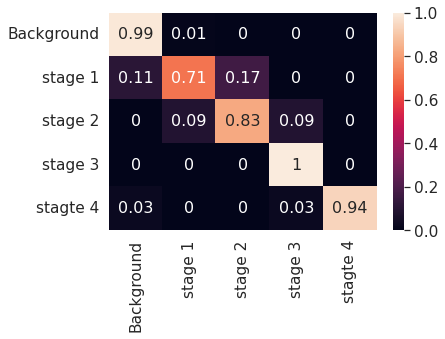}
        \caption{VGG-16 method}
        \label{fig:VGG-16R}
    \end{subfigure}

    \par\medskip

    \begin{subfigure}[b]{0.45\textwidth}
        \includegraphics[width=\textwidth]{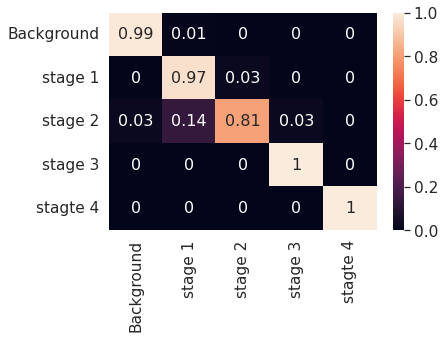}
        \caption{VGG-19 method}
        \label{fig:VGG-19R}
    \end{subfigure}
    \hspace{0.005\textwidth}
    \begin{subfigure}[b]{0.45\textwidth}
        \includegraphics[width=\textwidth]{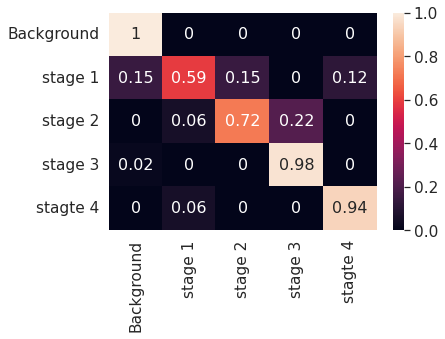}
        \caption{ResNet-50 method}
        \label{fig:ResNet50R}
    \end{subfigure}

    \caption{Confusion matrices for the canola test dataset. Rows represent true classes, columns represent predictions.}
    \label{fig:cm1}
\end{figure}

\begin{figure}[ht!]
    \centering
    \begin{subfigure}[b]{0.45\textwidth}
        \includegraphics[width=\textwidth]{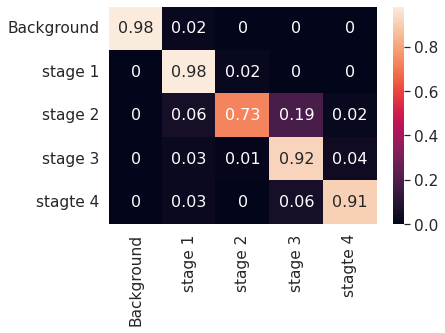}
        \caption{Gabor method}
        \label{fig:GR}
    \end{subfigure}
    \hspace{0.005\textwidth}
    \begin{subfigure}[b]{0.45\textwidth}
        \includegraphics[width=\textwidth]{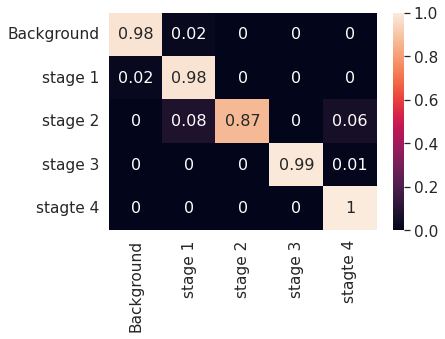}
        \caption{VGG-16 method}
        \label{fig:VR16}
    \end{subfigure}

    \par\medskip

    \begin{subfigure}[b]{0.45\textwidth}
        \includegraphics[width=\textwidth]{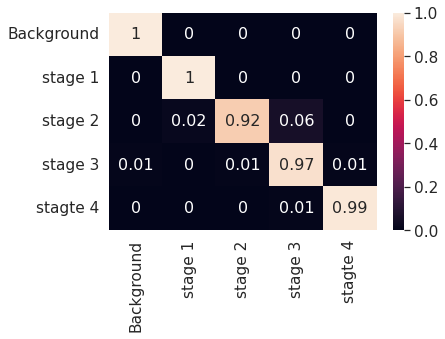}
        \caption{VGG-19 method}
        \label{fig:V19R}
    \end{subfigure}
    \hspace{0.005\textwidth}
    \begin{subfigure}[b]{0.45\textwidth}
        \includegraphics[width=\textwidth]{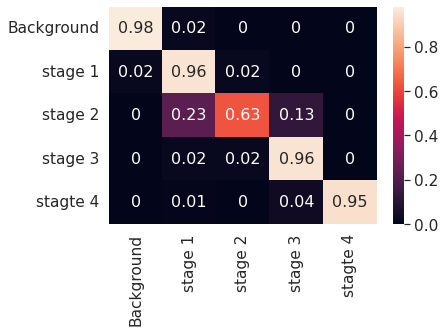}
        \caption{ResNet-50 method}
        \label{fig:R50R}
    \end{subfigure}

    \caption{Confusion matrices for the radish test dataset. Rows represent true classes, columns represent predictions.}
    \label{fig:cm2}
\end{figure}

Confusion matrices provide additional insight into classification performance. Figure~\ref{fig:cm1} and Figure~\ref{fig:cm2} summarise the results for both datasets. A notable pattern emerges: most models struggle with adjacent growth stages, which arise from gradual morphological transitions. For canola, misclassification occurs primarily between stages 1 and 2, where early stages share similar cotyledon structures and discriminative textures emerge only as true leaves develop. For radish, false negatives and false positives are frequent between stages 2 and 3. Transfer learning methods generally showed fewer classification errors compared to the Gabor method, indicating that this represents a fundamental visual limitation rather than a methodological weakness.
%
%

\subsection{Execution Time Analysis}
For real-time prediction, processing time is important. We tested execution time for all feature extraction methods using an HP ProBook 450 7th Generation laptop. Table~\ref{testing time} shows the average testing time per image for both datasets.

The Gabor filter-based models required more time per image than CNN-based models, with the best case (Gabor filter \& RBF-SVM) averaging 0.11-0.12 seconds per image. The most accurate model (VGG-19 \& RBF-SVM) achieves 0.08 seconds per image on both datasets. These prediction times were obtained without GPU acceleration, suggesting that both approaches are suitable for real-time implementation.
\begin{table}[ht!]
	\centering
	\caption{Testing times of all models on both datasets (seconds per image).}
        \label{testing time}
	\begin{tabular}{c c c}
		\hline
		\centering \textbf{Method} & \textbf{Canola} & \textbf{Radish} \\
		\hline
		VGG-16 \& RBF SVM & 0.07 & 0.06\\
		\hline
        VGG-16 \& XGBoost & 0.08 & 0.07\\
		\hline
		VGG-19 \& RBF SVM & 0.08 & 0.08\\
		\hline
        VGG-19 \& XGBoost & 0.09 & 0.09\\
		\hline
		ResNet-50 \& RBF SVM & 0.09 & 0.09\\
		\hline
        ResNet-50 \& XGBoost & 0.10 & 0.11\\
		\hline
		Gabor Filter \& RBF SVM & 0.12 & 0.11\\
		\hline
        Gabor Filter \& XGBoost & 0.30 & 0.31\\
		\hline
	\end{tabular} 
\end{table}
The Gabor filter-based method takes longer due to the morphological operations required before feature extraction, while CNN-based transfer learning models achieve faster processing since the convolutional layers used as feature extractors have been pre-trained.

\section{Conclusion}
\label{conclusion}
In this study, we applied two feature extraction methods to estimate plant growth stages: one that uses a bank of Gabor filters and morphological operators, and another based on transfer learning from pre-trained CNNs. We evaluated their performance experimentally using two custom datasets of two plant species derived from the publicly available "bccr-segset" dataset. We fine-tuned two classification algorithms, RBF-SVM and XGBoost, and concatenated them with feature extraction methods to create a complete growth stage classification pipeline.

We used PCA and t-SNE to visualise growth stage image separation in data sets, before and after feature extraction. Initially, clustering showed significant overlap among growth stages, highlighting similar plant features across classes and the need for effective feature extraction. The improved clustering after applying our feature extraction methods demonstrates their capability to extract features that enable accurate classification with minimal visible overlap between classes.
Experimental results showed that the CNN-based transfer learning method can achieve an average accuracy of 98.4\% in estimating four growth stages, while the Gabor method can achieve an average classification accuracy of 94.05\%. 
Using the same hardware setup, we also tested the execution time of both approaches. The VGG-16 SVM feature extractor had the fastest execution time per image, followed by the VGG-19 feature extractor. The Gabor filter SVM method had a higher execution time than the CNN-based methods due to the additional morphological operators. For real-time applications, both approaches demonstrate practical feasibility.
\subsection*{Acknowledgements}

This work was partially supported by the Telkom CoE at NWU. 
In addition, this work is based on the research supported in part by the National Research Foundation of South Africa (Grant Reference Numbers: RCDL240215206999).

 \bibliographystyle{IEEEtran}
 \bibliography{mybibliography}

\end{document}